\documentclass[journal]{IEEEtran}
\usepackage{csquotes}
\usepackage{lipsum}
\usepackage{url}
\usepackage{hyperref}

\usepackage{filecontents}
\usepackage{array}
\usepackage{multirow}
\usepackage{makecell}
\usepackage[square,sort,comma,numbers]{natbib}

\ifCLASSINFOpdf
  \usepackage[pdftex]{graphicx}
\else
\fi
\usepackage{algorithmic}

\usepackage{array}
\begin{document}
%
\title{Autonomous Vehicles: Evolution of Artificial Intelligence and Learning Algorithms}
%
%
%

\author{Divya Garikapati,~\IEEEmembership{Senior Member,~IEEE}, Sneha Sudhir Shetiya,~\IEEEmembership{Senior Member,~IEEE}

\thanks{Corresponding Author: Divya Garikapati is a Senior IEEE Member. E-mail: (divygari@umich.edu, divya.garikapati@ieee.org),  \\Sneha.S.Shetiya is a Senior IEEE Member. Email:(sneha.shetiya@ieee.org)}
\thanks{Manuscript created January 20, 2024.}}

\maketitle
\begin{abstract}
The advent of autonomous vehicles has heralded a transformative era in transportation, reshaping the landscape of mobility through cutting-edge technologies. Central to this evolution is the integration of Artificial Intelligence (AI) and learning algorithms, propelling vehicles into realms of unprecedented autonomy. This paper provides a comprehensive exploration of the evolutionary trajectory of AI within autonomous vehicles, tracing the journey from foundational principles to the most recent advancements.

Commencing with a current landscape overview, the paper delves into the fundamental role of AI in shaping the autonomous decision-making capabilities of vehicles. It elucidates the steps involved in the AI-powered development life cycle in vehicles, addressing ethical considerations and bias in AI-driven software development for autonomous vehicles. The study presents statistical insights into the usage and types of AI/learning algorithms over the years, showcasing the evolving research landscape within the automotive industry. Furthermore, the paper highlights the pivotal role of parameters in refining algorithms for both trucks and cars, facilitating vehicles to adapt, learn, and improve performance over time. It concludes by outlining different levels of autonomy, elucidating the nuanced usage of AI and learning algorithms, and automating key tasks at each level. Additionally, the document discusses the variation in software package sizes across different autonomy levels.

\end{abstract}

\begin{IEEEkeywords}
Artificial Intelligence (AI), Machine Learning (ML), Deep Neural Networks (DNNs), Natural Language Processing (NLP), Autonomous Vehicles (AVs), Safety, Security, Ethics, Emerging Trends, Trucks vs.Cars, Autonomy Levels, Operational Design Domain (ODD), Software-Defined Vehicles (SDVs), Connected and Automated Vehicles (CAVs), In-Vehicle AI Assistant, Internet Of Things (IOT), Natural Language Processing (NLP), Generative AI (GenAI).
\end{IEEEkeywords}

%
\IEEEpeerreviewmaketitle

\section{\textbf{Introduction}}
%
%
%
%
\IEEEPARstart{A}{rtificial} Intelligence (AI) and learning algorithms such as Machine Learning (ML), Deep Learning using Deep Neural Networks (DNNs) and Natural Language Processing (NLP) currently play a crucial role in the development and operation of autonomous vehicles. The integration of AI and learning algorithms enable autonomous vehicles to navigate, perceive, and adapt to dynamic environments, making them safer and more efficient. Continuous advancements in AI technologies are expected to further enhance the capabilities and safety of autonomous vehicles in the future.  Autonomous system development has been experiencing a transformational evolution through the integration of Artificial Intelligence (AI). This revolutionary combination holds the promise of reshaping traditional development processes, enhancing efficiency, and accelerating innovation. AI technologies are becoming integral in numerous facets of software development within autonomous vehicles making a paradigm shift towards Software-Defined Vehicles (SDVs) [\cite{1}][\cite{2}]. The success of autonomous vehicles hinges on balancing their potential benefits with addressing the challenges through collaborative efforts in technology development, regulation, and public communication. Some of the challenges include: 
\begin{itemize}
    \item \textit{Safety and Reliability:} Ensuring flawless AI performance in all scenarios is paramount.
    \item \textit{Regulations and Law:} Clear standards for safety, insurance, and liability are needed.
    \item \textit{Public Trust and Acceptance:}Addressing concerns about safety, data privacy, and ethical dilemmas is crucial.
    \item \textit{Cybersecurity: }Protecting against hacking and unauthorized access is essential.
    \item \textit{Ethical Dilemmas: }Defining AI decision-making in ambiguous situations raises moral questions.
    \item \textit{Addressing Edge cases: }Being able to handle unforeseen scenarios is challenging as those scenarios are rare and could be hard to imagine in some cases.
\end{itemize}

\subsection{\textbf{Benefits of AI/Learning Algorithms for Autonomous Vehicles}}

AI/Learning Algorithms are currently influencing various stages from initial coding to post-deployment maintenance in autonomous vehicles. Some of the benefits include:
\begin{itemize}
    \item \textbf{\textit{Safety:}} AI can significantly reduce accidents by eliminating human error, leading to safer roads.
    \item \textbf{\textit{Traffic Flow:}} Platooning and efficient routing can ease congestion and improve efficiency.
    \item \textbf{\textit{Accessibility:}} People with physical impairments or different abilities, the elderly, and the young can gain independent mobility.
    \item \textbf{\textit{Energy Savings:}} Optimized driving reduces fuel consumption and emissions.
    \item \textbf{\textit{Productivity and Convenience:}} Passengers use travel time productively while delivery services become more efficient.
\end{itemize}

AI in autonomous vehicles is poised for a bright future, shaping everyday life and creating exciting opportunities. Here's a glimpse of the possibilities:

\subsubsection{\textbf{\textit{Technological Advancements}}}
\begin{itemize}
    \item \textit{Sharper perception and decision-making:} AI algorithms are more adept at understanding environments with advanced sensors and robust machine learning.
    \item \textit{Faster, more autonomous operation:} Edge computing enables on-board AI processing for quicker decisions and greater independence.
    \item \textit{Enhanced safety and reliability:} Redundant systems and rigorous fail-safe mechanisms prioritizes safety above all else.
\end{itemize}

\subsubsection{\textbf{\textit{Education and Career Boom}}}
\begin{itemize}
    \item \textit{Surging demand for AI expertise:} Specialized courses and degrees in autonomous vehicle technology will cater to a growing need for AI, robotics, and self-driving car professionals.
    \item \textit{Interdisciplinary skills will be key: }Professionals with cross-functional skills bridging AI, robotics, and transportation will be highly sought after.
    \item \textit{New career paths in safety and ethics: }Expertise in ethical considerations, safety audits, and regulatory [\cite{3}] compliance will be crucial as self-driving cars become widespread.
\end{itemize}

\subsubsection{\textbf{Regulatory Landscape}}
\begin{itemize}
    \item \textit{Standardized safety guidelines: }Governments will establish common frameworks for performance and safety, building public trust and ensuring industry coherence.
    \item \textit{Stringent testing and validation:} Autonomous systems will undergo rigorous testing before deployment, guaranteeing reliability and safety standards.
    \item \textit{Data privacy and security safeguards:} Laws and regulations will address data privacy and cybersecurity concerns, protecting personal information and mitigating cyberattacks.
    \item \textit{Ethical and liability frameworks:} Clearly defined legal frameworks will address ethical decision-making and determine liability in situations involving self-driving cars.
\end{itemize}
This future holds immense potential for revolutionizing transportation, creating new jobs, and improving safety. However, navigating ethical dilemmas, ensuring robust regulations, and building public trust will be crucial for harnessing this technology responsibly and sustainably.

\subsection{\textbf{Operational Design Domains (ODDs) Expansions into new areas and Diversity - The Current Industry Landscape}}
These examples illustrate the diverse evolution of Operational Design Domains (ODDs) [\cite{4}] across various vehicle types, including trucks and cars, and within different geographical locations such as the United States, China, and Europe.
\begin{itemize}
    \item \textbf{\textit{Waymo Driver:}} Can handle a wider range of weather conditions, city streets, and highway driving, but speed limitations and geo-fencing restrictions apply.
    \item \textbf{\textit{Tesla Autopilot:}} Primarily for highway driving with lane markings, under driver supervision, and within specific speed ranges.
    \item \textbf{\textit{Mobileye Cruise AV:}} Operates in sunny and dry weather, on highways with clearly marked lanes, and at speeds below 45 mph.
    \item \textbf{\textit{Aurora and Waymo via: }}Wider range of weather conditions, including light rain/snow. Variable lighting (sunrise/sunset), Multi-lane highways and rural roads with good pavement quality, Daytime and nighttime operation, moderate traffic density, dynamic route planning,  Traffic light/stop sign recognition, intersection navigation, maneuvering in yards/warehouses etc.,
    \item  \textbf{\textit{TuSimple and Embark Trucks: }}Sunny, dry weather, clear visibility. Temperature range -10°C to 40°C, Limited-access highways with clearly marked lanes, Daytime operation only, maximum speed 70 mph, limited traffic density, pre-mapped routes, Lane changes, highway merging/exiting, platooning with other AV trucks etc.,
    \item \textbf{\textit{Pony.ai and Einride: }}Diverse weather conditions, including heavy rain/snow. Variable lighting and complex urban environments, Narrow city streets, residential areas, and parking lots, Low speeds (20-30 mph), high traffic density, frequent stops and turns, geo-fenced delivery zones,  Pedestrian and cyclist detection/avoidance, obstacle avoidance in tight spaces, dynamic rerouting due to congestion etc.,
    \item \textbf{\textit{Komatsu Autonomous Haul Trucks, Caterpillar MineStar Command for Haul Trucks: }} Harsh weather conditions (dust, heat, extreme temperatures). Limited or no network connectivity, Unpaved roads, uneven terrain, steep inclines/declines, Autonomous operation with remote monitoring, pre-programmed routes, high ground clearance, Obstacle detection in unstructured environments, path planning around natural hazards, dust/fog mitigation, etc.,
    \item \textbf{\textit{Baidu Apollo:}} Highways and city streets in specific zones like Beijing and Shenzhen. Operates in daytime and nighttime, under clear weather conditions, and limited traffic density. Designed for passenger transportation and robotaxis. Specific scenarios include Lane changes, highway merging/exiting, traffic light/stop sign recognition, intersection navigation, low-speed maneuvering in urban areas.
\end{itemize}
\begin{itemize}
    \item \textbf{\textit{WeRide:}} Limited-access highways and urban streets in Guangzhou and Nanjing. Operates in daytime and nighttime, under clear weather conditions. Targeted for robotaxi services and last-mile delivery. Specific scenarios include Lane changes, highway merging/exiting, traffic light/stop sign recognition, intersection navigation, automated pick-up and drop-off for passengers/packages.
\end{itemize}
\begin{itemize}
    \item \textbf{\textit{Bosch \& Daimler:}} Motorways and specific highways in Germany. Operates in daytime and nighttime, under good weather conditions. Focused on highway trucking applications. Specific scenarios include Platooning with other AV trucks, automated lane changes and overtaking, emergency stopping procedures, communication with traffic management systems.
\end{itemize}
\begin{itemize}
    \item \textbf{\textit{Volvo Trucks:}} Defined sections of Swedish highways. Operates in daytime and nighttime, under varying weather conditions. Tailored for autonomous mining and quarry operations. Specific scenarios include Obstacle detection and avoidance in unstructured environments, path planning around natural hazards, pre-programmed routes with high precision, remote monitoring and control.
\end{itemize}

      In this paper, we discuss the AI-powered software development lifecycle for autonomous vehicles and discuss the details on how to ensure software quality, security and resolve ethical dilemmas by taking different biases into account during the development of the AI algorithms. We explain how the AI algorithms have been emerging and evolving over time to have more and more decision making capabilities without human involvement using  IOT as a future direction of expansion for autonomous vehicles to being more connected to other actors in the driving environment. In-cabin experience enhancements and Driver Assistant Systems were also discussed as part of the emerging trends. A literature survey of how the AI algorithms are being used within autonomous vehicles has been provided in Section V.  In Section VI, we have also provided certain statistics on how the use of AI and Learning algorithms have been evolving over time, how the research in these areas has been trending over time, different AI model parameters being considered for autonomous trucks vs. passenger cars etc.,. Another interesting statistic on how the use of AI and Learning algorithms change based on the levels of autonomy was also provided.

\section{The AI-powered Development Life-Cycle in Autonomous Vehicles}
This section describes about the key aspects involved with the AI-powered development life cycles within autonomous vehicles and these could be applicable to other fields as well in general.

\subsection{\textbf{Model Training and Deployment}}
AI model training and deployment in autonomous vehicles involves a systematic process and typically includes several stages:

\textit{Data Collection and Pre-processing:} Gathering a vast amount of data from real-world sensors, pre-existing datasets and other sources such as synthetic datasets. Cleaning and pre-processing the data to make it suitable for machine learning models.

\textit{Model Training:} Employ learning models such as neural networks, deep learning [\cite{5}], or natural language processing (NLP) to understand patterns and structures based on the data. Training the models to a desired level of accuracy based on each scenario or in generic abstract cases like being able to extract the patterns during the live operation of the vehicles.

\textit{Model Generation:} Train models to perform a certain decision making task, functions, or modules based on learned patterns. These models can use various architectures, such as decision trees, random forests, regression trees, deep layers, ensemble learning etc., 

\textit{Code Refinement and Optimization[\cite{6}]:} Refine the generated code to improve its quality, readability, and functionality. Post-generation processing ensures the code adheres to coding standards, conventions [\cite{7}] and requirements.

\textit{Quality Assessment:} Evaluate the generated code for correctness, efficiency, and adherence to the intended functionalities. This involves testing, debugging, and validation procedures.

\textit{Integration and Deployment:} Integrate the model into the broader system under development for autonomy implementation. Deploy and test the software application incorporating the new model using multiple methods like software-in-the-loop, hardware-in-the-loop, human-in-the-loop etc., using simulation, closed course and limited public road environments. Some models are trained to improve their learning even after deployment. These models need to be tested for future directions of learning to ensure compliance to ethical considerations as explained in Section III and other requirements.

Using a systematic process like this would help build the confidence levels on each model being developed and deployed in various subsystems of autonomous vehicles like perception, planning, controls and Human-Machine Interface (HMI) applications.
\begin{figure}
    \centering
    \includegraphics[scale=0.10]{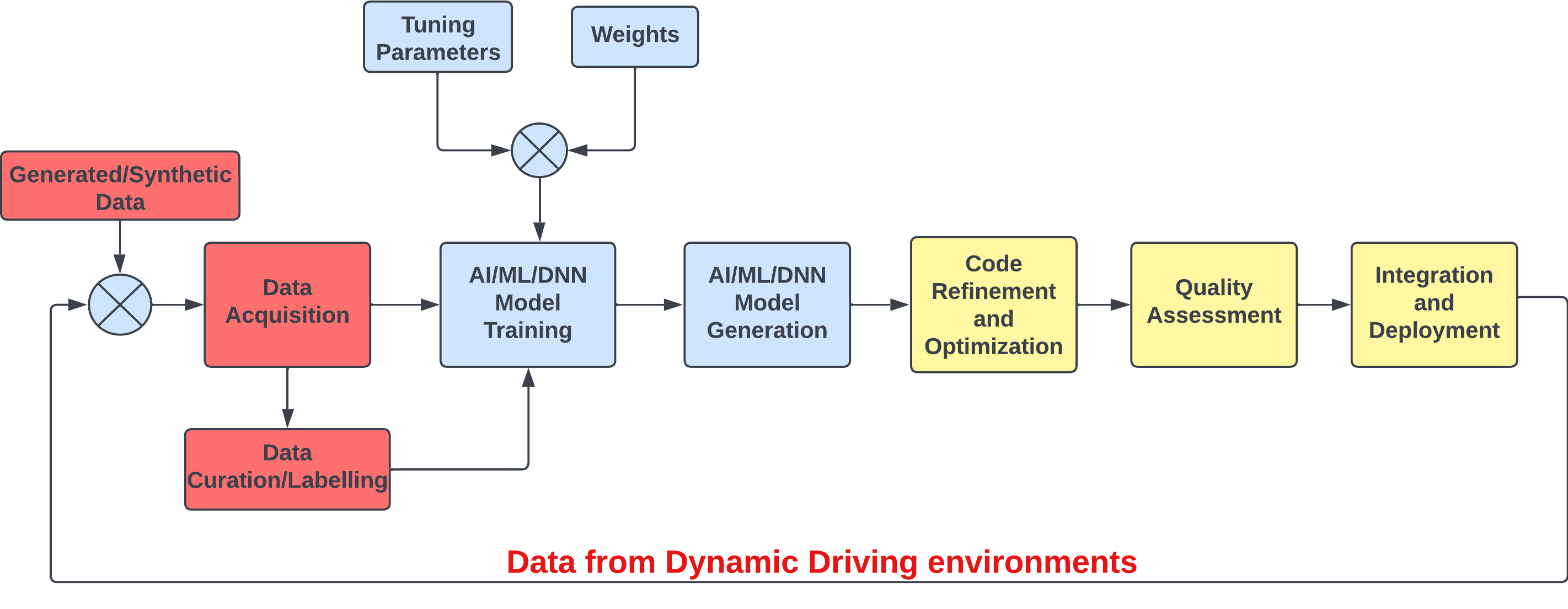}
    \caption{AI-Powered Development Life-Cycle}
    \label{fig:AI-devcycle}
\end{figure}

\subsection{\textbf{Ensuring Software Quality and Security}}
In autonomous vehicles, the integration of AI in various aspects of software development and maintenance plays a crucial role in ensuring the robustness and security of the overall system. Automated testing, powered by AI-based tools, emerges as a key component in the testing process. These tools efficiently identify bugs, vulnerabilities, and ensure that the software functions as intended, contributing to the reliability of autonomous vehicle software. Additionally, AI extends its capabilities to code analysis and review, providing a thorough examination of the codebase for quality and highlighting potential issues or vulnerabilities. Predictive maintenance, facilitated by AI, becomes essential for anticipating and addressing potential software failures, ultimately reducing downtime and enhancing the overall operational efficiency of autonomous vehicles. Moreover, AI-driven anomaly detection and security monitoring contribute significantly to the safety of autonomous vehicles. By continuously monitoring the software environment, AI systems can identify abnormal patterns or behaviors, promptly responding to potential security threats in real time. Vulnerability assessment, another application of AI tools, conducts in-depth evaluations to pinpoint weaknesses in software systems, providing valuable insights to mitigate risks effectively.
Behavioral analysis powered by AI proves instrumental in understanding user interactions within the software. This capability aids in detecting and preventing suspicious or malicious activities, fostering a secure and reliable autonomous vehicle ecosystem. Finally, AI's role in fraud detection within software applications adds an extra layer of security, ensuring the integrity of autonomous vehicle systems and safeguarding against potential security breaches. In summary, the integration of AI in these diverse areas significantly enhances the overall safety, security, and efficiency of autonomous vehicles.

\section{Ethical Considerations and Bias in AI-Driven Software Development for Autonomous vehicles}
To address the challenges related to bias, understanding and addressing these concerns are crucial for building responsible and fair AI-driven software for autonomous vehicles. Here are key points highlighting ethical considerations and bias in AI-driven software development:

\begin{enumerate}
    \item \textbf{Data Bias:}
\begin{itemize}
        \item Challenge: AI models learn from historical data, and if the training data is biased, the model can perpetuate and amplify existing biases.
        \item Mitigation: Rigorous data pre-processing, diversity in training data, and continuous monitoring for bias are essential. Ethical data collection practices must be upheld.
\end{itemize}

    \item \textbf{Algorithmic Bias:}

\begin{itemize}
        \item Challenge: Algorithms may inadvertently encode biases present in the training data, leading to discriminatory outcomes.
        \item Mitigation: Regular audits of algorithms for bias, transparency in algorithmic decision-making, and the incorporation of fairness metrics during model evaluation.
\end{itemize}

    \item \textbf{Fairness and Accountability:}

\begin{itemize}
        \item Challenge: Ensuring fair outcomes and establishing accountability for AI decisions is complex, especially when models are opaque.
        \item Mitigation: Implementing explainable AI (XAI) techniques, defining clear decision boundaries, and establishing accountability frameworks for AI-generated decisions.
\end{itemize}

    \item \textbf{Explainability and Transparency:}

\begin{itemize}
        \item Challenge: Many AI models operate as "black boxes," making it challenging to understand how decisions are reached. AI safety is another challenge that needs to be made sure is safety-critical applications like autonomous vehicles[9]
        \item Mitigation: Prioritizing explainability [\cite{8}] in AI models, using interpretable algorithms, and providing clear documentation on model behavior.
\end{itemize}

    \item \textbf{User Privacy:}

\begin{itemize}
        \item Challenge: AI systems often process vast amounts of personal data, raising concerns about user privacy.
        \item Mitigation: Implementing privacy-preserving techniques, obtaining informed consent, and adhering to data protection regulations (e.g., GDPR [\cite{9}]) to safeguard user privacy.
\end{itemize}

    \item \textbf{Security Concerns:}

\begin{itemize}
        \item Challenge: AI models can be vulnerable to adversarial attacks, posing security risks.
        \item Mitigation: Robust testing against adversarial scenarios, incorporating security measures, and regular updates to address emerging threats.
\end{itemize}

    \item \textbf{Inclusivity and Accessibility:}

\begin{itemize}
        \item Challenge: Biases in AI can result in excluding certain demographics, reinforcing digital divides.
        \item Mitigation: Prioritizing diversity in development teams, actively seeking user feedback, and conducting accessibility assessments to ensure inclusivity.
\end{itemize}

    \item \textbf{Social Impact:}

\begin{itemize}
        \item Challenge: The deployment of biased AI systems can have negative social implications, affecting marginalized communities disproportionately.
        \item Mitigation: Conducting thorough impact assessments, involving diverse stakeholders in the development process, and considering societal consequences during AI development.
\end{itemize}

    \item \textbf{Continuous Monitoring and Adaptation:}

\begin{itemize}
        \item Challenge: AI models may encounter new biases or ethical challenges as they operate in dynamic environments.
        \item Mitigation: Establishing mechanisms for ongoing monitoring, feedback loops, and model adaptation to address evolving ethical considerations.
\end{itemize}

    \item \textbf{Ethical Frameworks and Guidelines:}

\begin{itemize}
        \item Challenge: The absence of standardized ethical frameworks can lead to inconsistent practices in AI development.
        \item Mitigation: Adhering to established ethical guidelines, such as those provided by organizations like the ISO, IEEE, SAE, Government regulatory boards etc., and actively participating in the development of industry-wide standards.
\end{itemize}

\end{enumerate}
Addressing ethical considerations and bias in AI-driven software development in autonomous vehicles requires a holistic and proactive approach[\cite{10}]. It involves a commitment to fairness, transparency, user privacy, and social responsibility throughout the AI development lifecycle. As the field evolves, continuous efforts are needed to refine ethical practices and promote responsible AI deployment.

\section{AI's role in the Emerging trend of Internet of Things (IoT) Ecosystem for Autonomous vehicles}
Artificial Intelligence (AI) plays a crucial role in shaping and enhancing the capabilities of the Internet of Things (IoT). Here's an overview of how AI contributes to the IoT Ecosystem for Autonomous vehicles

\begin{figure}
    \centering
    \includegraphics[scale=0.05]{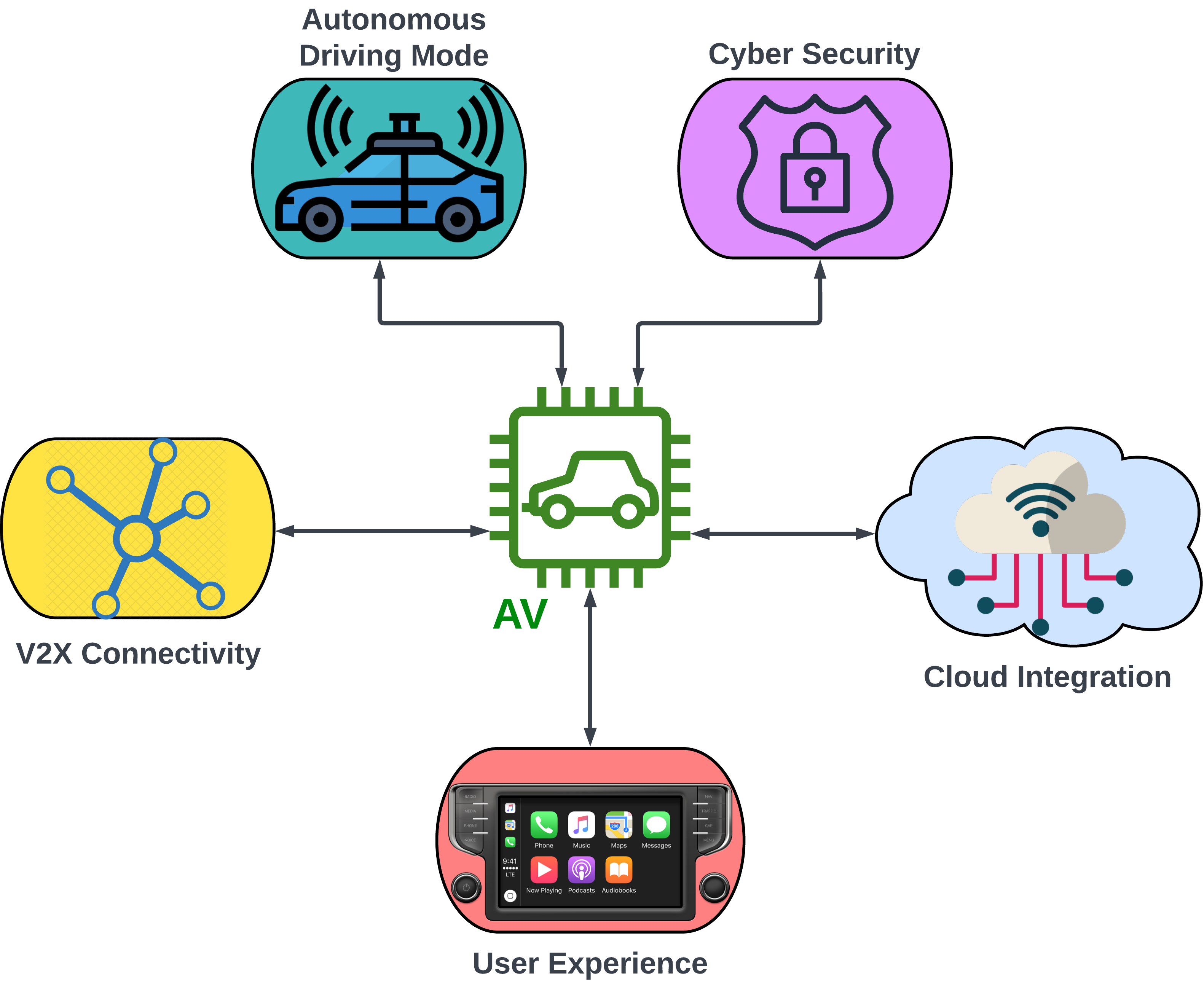}
    \caption{AI's role in the Internet of Things (IOT) Ecosystem for Autonomous vehicles}
    \label{fig:IOT-for-AVs}
\end{figure}

In the realm of Connected and Autonomous Vehicles (CAVs), AI and IoT converge to create a seamless network of intelligence and connectivity, transforming the driving experience. Vehicles become intelligent agents, processing sensor data in real-time to make informed decisions: predicting traffic patterns, optimizing routes, detecting anomalies, and even adapting to changing road conditions with dynamic adjustments. This intelligent ecosystem extends beyond individual vehicles, interconnecting with infrastructure and other vehicles to optimize traffic flow, anticipate potential hazards, and personalize the driving experience.
 
 Key AI-powered IoT capabilities in CAVs include the following:
\begin{itemize}
    \item Real-time data processing and analysis for insights into traffic, road conditions, and vehicle health.
    \item Predictive analytics for proactive maintenance, efficient resource allocation, and informed decision-making.
    \item Enhanced automation for autonomous driving tasks, adaptive cruise control, and dynamic route optimization.
    \item Efficient resource management for optimizing energy consumption, bandwidth usage, and load balancing.
    \item Security and anomaly detection for identifying potential threats and preventing cyberattacks.
    \item Personalized user experience through customized settings, preferences, and tailored insights.
    \item Edge computing for real-time decision-making, reducing latency and improving responsiveness.
\end{itemize}
Challenges to address include ensuring data privacy, security, interoperability, and overcoming resource constraints in connected vehicles. The seamless integration of AI and IoT holds the potential to revolutionize transportation, leading to safer, more efficient, and sustainable [\cite{11}] mobility solutions.

\subsection{\textbf{Enhancing User Experience}}

\textit{\textbf{\textit{Personalization and Recommendation Systems in-cabin:}} }AI-driven personalization and recommendation systems in Autonomous vehicles use machine learning models to analyze user behavior and preferences, creating personalized recommendations for tools, libraries, and vehicle maneuvers. They collect and pre-process user data, create individual profiles, generate tailored suggestions, and continuously adapt based on real-time interactions, aiming to enhance user experience and developer productivity.

\textbf{\textit{Natural Language Processing (NLP) in-cabin:}} NLP enables software to comprehend and process human language. This includes chat bots [\cite{12}], virtual assistants and voice recognition systems that understand and respond to natural language queries in vehicle cabins. It allows the vehicle subsystems to analyze and derive insights from user requirements and structuring requirements effectively to create responses and certain vehicle maneuvers. 

\textbf{\textit{Generative Artificial Intelligence (Gen AI):}} This technology uses machine learning algorithms to produce new and original outputs based on the patterns and information it has learned from training data. In the context of vehicles, generative AI can be applied to various aspects, including natural language processing for in-car voice assistants, content generation for infotainment systems, and even simulation scenarios for testing autonomous driving systems. Large Language Models (LLMs) are a specific class of generative AI models that are trained on massive amounts of text data to understand and generate human-like language.In vehicles, LLMs can be employed for natural language understanding and generation, allowing for more intuitive and context-aware interactions between the vehicle and its occupants. This can enhance features like voice-activated controls, virtual assistants, and communication systems within the vehicle.

\section{Review of Existing Research and Use Cases}

\textbf{H. J. Vishnukumar et. al. [\cite{12}]} introduced that traditional development methods like Waterfall and Agile, fall short when testing intricate autonomous vehicles and proposes a novel AI-powered methodology for both lab and real-world testing and validation (T\&V) of ADAS and autonomous systems. Leveraging machine learning and deep neural networks, the AI core learns from existing test scenarios, generates new efficient cases, and controls diverse simulated environments for exhaustive testing. Critical tests then translate to real-world validation with automated vehicles in controlled settings. Constant learning from each test iteration refines future testing, ultimately saving precious development time and boosting the efficiency and quality of autonomous systems. This methodology lays the groundwork for AI to eventually handle most T\&V tasks, paving the way for safer and more reliable autonomous vehicles.

\textbf{Bachute, Mrinal R et. al. [\cite{13}]} described the algorithms crucial for various tasks in Autonomous Driving, recognizing the multifaceted nature of the system. It discerns specific algorithmic preferences for tasks, such as employing Reinforcement Learning (RL) models for effective velocity control in car-following scenarios and utilizing the "Locally Decorrelated Channel Features (LDCF)" algorithm for superior pedestrian detection. The study emphasizes the significance of algorithmic choices in motion planning, fault diagnosis with data imbalance, vehicle platoon scenarios, and more. Notably, it advocates for the continuous optimization and expansion of algorithms to address the evolving challenges in Autonomous Driving. The paper serves as an insightful foundation, prompting future research endeavors to broaden the scope of tasks, explore a diverse array of algorithms, and fine-tune their application in specific areas of interest within the Autonomous Driving System.

\textbf{Y. Ma et. al. [\cite{14}]} explained the pivotal role of artificial intelligence (AI) in propelling the development and deployment of autonomous vehicles (AVs) within the transportation sector. Fueled by extensive data from diverse sensors and robust computing resources, AI has become integral for AVs to perceive their environment and make informed decisions while in motion. While existing research has explored various facets of AI application in AV development, this paper addresses a gap in the literature by presenting a comprehensive survey of key studies in this domain. The primary focus is on analyzing how AI is employed in supporting crucial applications in AVs: 1) perception, 2) localization and mapping, and 3) decision-making. The paper scrutinizes current practices to elucidate the utilization of AI, delineating associated challenges and issues. Furthermore, it offers insights into potential opportunities by examining the integration of AI with emerging technologies such as high-definition maps, big data, high-performance computing, augmented reality (AR) and virtual reality (VR) enhanced simulation platforms, and 5G communication for connected AVs. In essence, this paper serves as a valuable reference for researchers seeking a deeper understanding of AI's role in AV research, providing a comprehensive overview of current practices and paving the way for future opportunities and advancements. \\

\section{AI and Learning Algorithms statistics for Autonomous vehicles}
This Section extends the analysis of Artificial Intelligence (AI) and Learning Algorithms in autonomous vehicles, building upon previous work as described in Section V. The focus is on providing additional statistical insights into the following:
- evolution of different types of AI and learning algorithms over the years, 
- research trends in application of AI in all fields vs. autonomous vehicles, 
- creation of a parameter set crucial for autonomous trucks versus cars, 
- evolution of AI and learning algorithms at different autonomy levels, and 
- changes in the types of algorithms, software package size etc., over time.

\subsection{\textbf{Stat1: Trends of usage of AI, ML and DNN Algorithms over the years}}
Today, a vehicle’s main goal is not limited to transportation, but also includes comfort, safety, and convenience. This led to extensive research on improving vehicles and incorporating technological breakthroughs and advancements.

As per prior work done for the development for architecture and ADAS technology  it is evident that the research till now has limitations. These limitations are pertaining either to the author’s elaboration of his/her knowledge or not having proper sources. Thus its a good exercise to have a look at the trends over the years as our capabilities to develop these ML models have gotten better and also the access to better computing units[\cite{15}][\cite{16}] has led to the evolution of the algorithms. 
In the Table \ref{tab:Stat1_table} , we have summarized different modelling algorithms for various standard components of the ADAS algorithm. The second column illustrates the technology that exists in today's date and the third column predicts potential future development which is efficient than the current.

\begin{table}[h!]
    \caption{Autonomous Driving: Key technologies evolution}
    \centering
    \fontsize{6.5pt}{7.5pt}\selectfont
    \begin{tabular}{|b{1.5cm}|b{2.5cm}|b{3.0cm}|} \hline 
        \textbf{Technology}&  \textbf{Developed over years}& \textbf{Future} \\ \hline 
         Environmental Perception& DL for object detection
         YOLOv3 
         K-means clustering& Very challenging. Needs more research to better detect objects in blurry, extreme and rare conditions in real time.\\ \hline 
        Pedestrian detection & PVANET and RCNN model for object detection during blurry weather  & OrientNet, RPN, and PredictorNet to solve occlusion problem \\ \hline 
        Path Planning & DL algorithm based on CNN & multisensor fusion system,
along with an INS, a GNSS, and a LiDAR system, would be used to implement a 3D SLAM.
\\ \hline 
        Vehicle Cybersecurity & Security testing and TARA & Remote control of AV deploying IoT sensors \\ \hline 
        Motion Planning & Hidden Markov model 
Q-learning algorithm
 & Grey prediction model utilising and Advanced model predictive control for effective lane change\\ \hline
    \end{tabular}
    \label{tab:Stat1_table}
\end{table}

Below we have derived series of plots pertaining to research publications in AI(Artificial Intelligence), ML(Machine Learning) and DNN(Deep Neural Network) domains. Brief explanations have been provided before to understand what topics come under these domains.

\paragraph{\textbf{AI (Artificial Intelligence)} }

\begin{itemize}
    \item Expert Systems: Rule-based systems that mimic human expertise for decision-making [\cite{17}].
    \item Decision Trees: Hierarchical structures for classification and prediction.ex: prognostics area
    \item Search Algorithms: Methods for finding optimal paths or solutions, such as A* search and path planning algorithms.
    \item Generative AI: To create scenarios for training the system and for balancing data on high severity accident/non-accident cases. (CRSS dataset). Create a non-existent scenario dataset. Supplement the real datasets. Simulation testing. 
    \item NLP: AI Assistant (Yui, Concierge, Hey Mercedes, etc.,) - LLMs
\end{itemize}

\paragraph{\textbf{ML (Machine Learning)}}
\begin{itemize}
    \item Supervised Learning: Algorithms that learn from labeled data to make predictions, such as:

\begin{itemize}
        \item Linear Regression: For predicting continuous values.
        \item Support Vector Machines (SVMs): For classification and outlier detection.
        \item Decision Trees: For classification and rule generation.
        \item Random Forests: Ensembles of decision trees for improved accuracy.
\end{itemize}

    \item Unsupervised Learning: Algorithms that find patterns in unlabeled data, such as:

\begin{itemize}
        \item Clustering Algorithms (K-means, Hierarchical): For grouping similar data points.
        \item Dimensionality Reduction (PCA, t-SNE): For reducing data complexity.
\end{itemize}

\end{itemize}

\paragraph{\textbf{DNN (Deep Neural Networks)}}

\begin{itemize}
    \item Convolutional Neural Networks (CNNs): For image and video processing, used for object detection, lane segmentation, and traffic sign recognition.
    \item Recurrent Neural Networks (RNNs): For sequential data processing, used for trajectory prediction and behavior modeling.
    \item Deep Reinforcement Learning (DRL): For learning through trial and error, used for control optimization and decision-making.
\end{itemize}

 \paragraph{\textbf{Specific Examples in Autonomous Vehicles}}

\begin{itemize}
    \item Object Detection (DNN): CNNs like YOLO [\cite{18}], SSD [\cite{19}], and Faster R-CNN  are used to detect objects around the vehicle.
    \item Lane Detection (DNN): CNNs are used to identify lane markings and road boundaries .
    \item Path Planning (AI): Search algorithms like A* and RRT are used to plan safe and efficient routes.
    \item Motion Control (ML): Regression models [\cite{20}][\cite{21}][\cite{22}] are used to predict vehicle dynamics and control steering, acceleration, and braking.
    \item Behavior Prediction (ML): SVMs or RNNs are used to anticipate the behavior of other vehicles and pedestrians.
\end{itemize}

\begin{figure}[h!]
    \centering
    \includegraphics[scale=0.35]{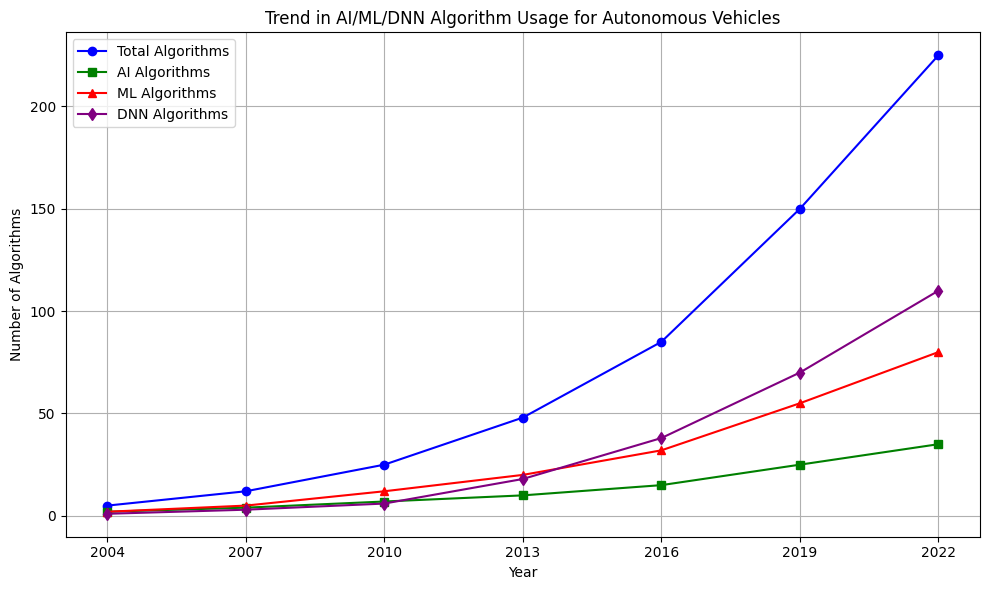}
    \caption{Trends of usage of AI,ML and DNN algorithms over the years}
    \label{fig:AI_Trend_1}
\end{figure}

In the Figure \ref{fig:AI_Trend_1} , we evaluated the papers from [\cite{23}][\cite{24}]and found the trends to be as shown. One can observe that year 2013 the number of algorithms in DNN surpasses that in generic AI and ML. This shows more research with deep neural networks and the traction it received in the AI community. However the main takeaway from the graph is the exponential upward trend in the number of algorithms over the years developed for AI applications.

Some research was done considering platforms of IEEEXplore, SAE Mobilus, MDPI and Science Direct to find out the published research in AI/ML and also particularly in Autonomous vehicles.

When filtering the MDPI Journals and articles, one can observe that there is an additional filter relating to \textit{Data} that pops up after 2021. This indicates that pre-2020 not much papers related to data handling and analysis were published as the collected data was not huge. One also observes that the year 2020 (year of the COVID pandemic) for MDPI; saw minimal papers for autonomous technology. While several factors may contribute to the rise in model deployments observed in 2021, a possible explanation is the limited opportunity for previous models to undergo real-world testing through vehicle deployment. Notably, the number of deployed models surged to 737 in 2021, representing nearly a twofold increase compared to earlier years. 

From the IEEE publications, one can see that although effective research in AI/ML increases over time not much research has been published towards autonomous vehicle technology.

Shifting Trends in IEEE Publications: Interestingly, post-2021, the upward trend in LMM and DNN publications (identified through filters aligned with our previous analysis) appears to plateau. This suggests a potential shift in research focus within computer vision (CV) following the emergence of Generative AI (GenAI) and other advanced technologies. While LMM and DNN remain foundational, their prominence as primary research subjects within classic CV might be declining. 

Considering CVPR Publications: Initially, we considered including CVPR publications in our analysis. However, we ultimately excluded them due to significant overlap with the IEEE dataset. As a significant portion of CVPR papers are subsequently published in IEEE journals, including both sets would introduce redundancy and potentially skew the analysis. 

\begin{figure}[h!]
    \centering
    \includegraphics[scale=0.45]{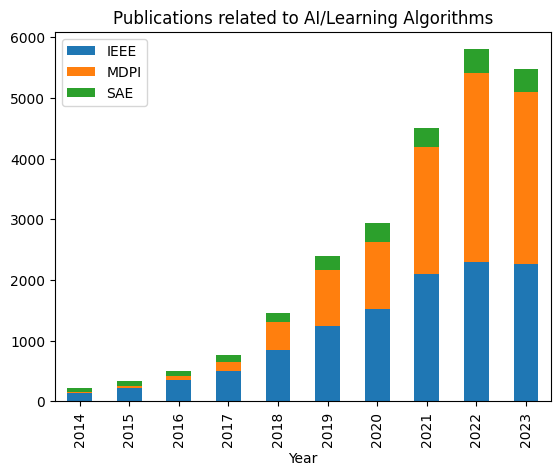}
    \caption{No. of Publications related to AI/learning Algorithms in all Fields}
    \label{fig:AI_Trend_2}
\end{figure}

\begin{figure}[h!]
    \centering
    \includegraphics[scale=0.45]{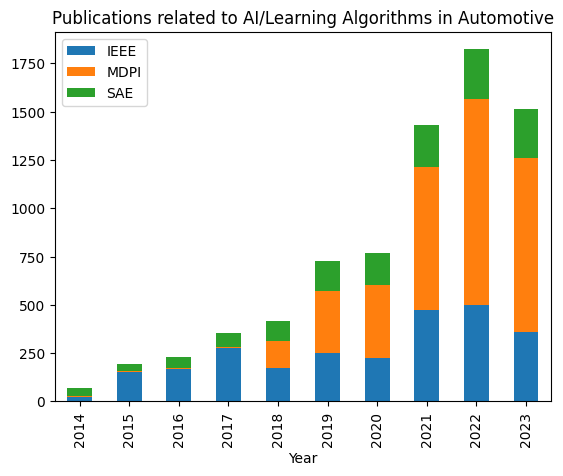}
    \caption{No. of Publications related to AI/learning Algorithms for Autonomous Vehicles}
    \label{fig:AI_Trend_3}
\end{figure}

\begin{figure}[h!]
    \centering
    \includegraphics[scale=0.45]{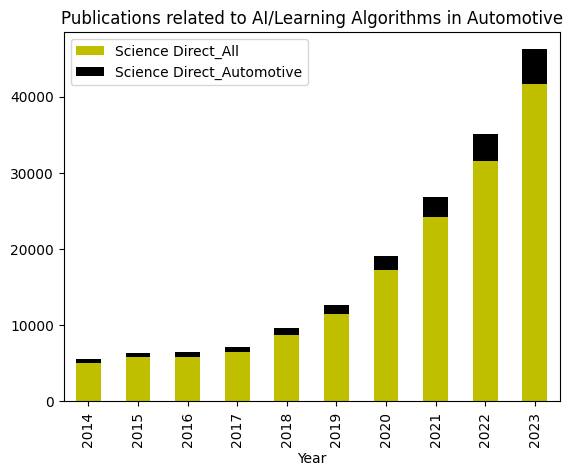}
    \caption{No. of Publications related to AI/learning Algorithms for Autonomous Vehicles vs. all Fields in Science Direct}
    \label{fig:AI_Trend_4}
\end{figure}

Figure \ref {fig:AI_Trend_2} focuses on all of AI/ML publications related to IEEEXplore [\cite{25}], MDPI (Multidisciplinary Digital Publishing Institute) [\cite{26}] and SAE (Society Of Automotive Engineers) [\cite{27}]
Figure  \ref{fig:AI_Trend_3} focuses on the trend changes in publications for autonomous vehicles.
Figure\ref{fig:AI_Trend_4} focuses on Science Direct [\cite{28}] where we see the publications are in thousands with very little presence for autonomous vehicles. This is an indication of how AI/ML applications have surpassed engineering and are used everywhere from medical to defence.

From the graphs we see comparatively less publications in starting years 2014-2018. There is a huge surge in 2018 where also we see Autonomous vehicles with advanced self driving features gained traction. From the trend, we expect in future a similar exponential rise. However we do expect additional parameters(for ex:data got introduced) to be in the list. With AI/ML applications coming up in every industry along with automotive, the future for research in the area is promising.
 

\subsection{\textbf{Stat2: Parameters for AI model (Trucks vs. Cars)}}

As per American Trucking Associations (ATA), there will be a shortage of over 100,000 truck drivers in the US by 2030, which could potentially double by 2050 if trends continue.
Bureau of Labor Statistics (BLS stats), while not explicitly predicting a shortage, the BLS projects a slower-than-average job growth for truck drivers through 2030, indicating potential challenges in meeting future demand. Aging work force, demanding job conditions and regulatory hurdles are few of the reasons which contribute towards the same. 

The above two results give a good business case for driverless trucks in comparison to driverless cars. This is also contradictory to the belief that truck drivers may loose jobs over the self-driving technology. In fact as per [\cite{29}],[\cite{30}],driverless trucks can drastically reduce the driver costs, increase truck utilization and improve truck safety. Inspite of this, one can see not enough research has been done on the impacts of self-driving trucks compared to passenger transport [\cite{31}]. There is a need to ensure road freight transport has alignment with its current operations retaining its value chain. One cannot think of cost reductions by taking out the driver cabin as most self-driving technology developing trucking companies are focusing on hub to hub transport and unlike passenger cars, not from source to destination. One would still need a driver in the start and end of the journey. This refocuses on the statement above for the need of truck drivers in future but eliminating the other drawbacks of long haul freight transport.

\begin{table}[htp]
    \centering
     \fontsize{6.5pt}{7.5pt}\selectfont
\caption{Parameter set for differences in using AI in autonomous trucks and cars}
\label{tab:parameter_set}
    \begin{tabular}{|p{1.1cm}|p{1.3cm}|p{2.4cm}|p{2.4cm}|} \hline 
         \textbf{Parameter}&  \textbf{ Sub-class}&\textbf{Trucks}& \textbf{Cars}\\ \hline 
         Environment&   Traffic density&operate on highways with predictable traffic patterns,& encounter diverse, often congested, urban environments.\\ \hline 
         &   Road infrastructure&navigate primarily on well-maintained highways&  deal with varied road conditions and potentially unmarked streets.\\ \hline 
         &   Weather conditions&may prioritize stability and visibility for cargo safety& may prioritize maneuverability for passenger comfort.\\ \hline 
         Vehicle characteristics&   Size and weight&Larger size and weight present different sensor ranges and dynamic response complexities& Smaller size and weight in comparison to Trucks.\\ \hline 
 & Cargo handling and safety& require AI to manage cargo weight distribution and potential shifting cargo&This is not a concern for cars\\ \hline 
 & Fuel efficiency and emissions& Truck AI prioritizes efficient fuel consumption due to long-distance travel&Car AI may prioritize smoother acceleration and deceleration for passenger comfort.\\ \hline 
 Operational considerations:& Route planning and optimization& require long-distance route planning with considerations for infrastructure limitations, rest stops, and cargo delivery schedules.&generally focus on shorter, dynamic routes with real-time traffic updates.\\ \hline 
 & Communication and connectivity&  may rely on dedicated infrastructure for communication (platooning, V2X)&primarily use existing cellular networks.\\\hline
         &   Legal and regulatory landscape&Regulations regarding automation and liability are tight.& regulations impacting AI and deployment are different than trucks.\\ \hline
 AI algorithm and hardware needs& Perception and sensor fusion& may prioritize radar and LiDAR for long-range detection& benefit from high-resolution cameras for near-field obstacle avoidance.\\ \hline 
 & Decision-making and planning& AI focuses on safe, fuel-efficient navigation and traffic flow optimization& AI prioritizes dynamic route adjustments, pedestrian/cyclist detection, and passenger comfort.\\ \hline 
 & Redundancy and safety protocols& may have stricter fail-safe measures due to cargo risks.&have safety protocols with redundant systems\\ \hline 
 Additional factors& Public perception and acceptance& Public trust in truck automation might be slower to build due to size and potential cargo risks.&Public trust in car automation is higher due to lesser risks\\ \hline 
 & Economic and business models& automation models may involve fleet management and logistics optimizations& automation may focus on ride-sharing and individual ownership\\\hline
    \end{tabular}

\end{table}

As mentioned in [\cite{30}] above, according to Daimler Ex-CEO Zetsche, future vehicles need to have four characteristics; connected, autonomous, shared, and electric, a so-called CASE vehicle. Nevertheless, each of these points has the potential to turn the industry upside down. The paper clearly states backed up by a study that level 4 automation will be reached by 2030 followed by level 5 in 2040. Based on the interview results conducted in [\cite{29}] and the delphi-based scenario study with projections for the next 10 years, it is evident one needs to seriously consider the impact of automation on trucks. Lots of research revolves around passenger cars with many competitors in the market. We found that not much data exists for self- driving for trucks.

This parameter set as shown in Table \ref{tab:parameter_set} serves as a starting point for understanding the key differences in how AI is applied to autonomous trucks and cars. Each parameter can be further explored and nuanced based on specific scenarios and applications. Currently, the autonomous trucking has been expanding in 4 major categories such as Highway Trucking ODD, Regional Delivery ODD, Urban logistics ODD and Mining and Off-Road ODD. There are 3 different categories as well based on the different stages of logistics to handle the movement of goods for autonomous trucking like Long Haul, Middle Mile and Last Mile. Understanding these categorization and how the trucking industry has been evolving to deliver more autonomous vehicles is really important for the future of logistics to help optimize and streamline the entire supply chain, ensuring efficient and timely delivery of goods to their final destination.

\subsection{\textbf{Stat3: Usage of AI and Learning Algorithms at various Levels of Autonomy}}

Autonomous vehicles operate at various levels of autonomy, from Level 0 to Level 5, each presenting unique challenges and opportunities. This section explores the diversity and evolution of AI algorithms across different levels of autonomous vehicle capabilities. Autonomous vehicles are categorized into different levels based on their autonomy, with increasing complexity and diversity of AI algorithms as autonomy levels progress. The six levels [\cite{34}] of AV autonomy define the degree of driver involvement and vehicle automation. At lower levels (L0-L2), driver assistance systems primarily utilize rule-based and probabilistic methods for specific tasks like adaptive cruise control or lane departure warning. Higher levels (L3-L4) rely heavily on machine learning and deep learning algorithms, particularly for perception tasks like object detection and classification using convolutional neural networks (CNNs). Advanced sensor fusion techniques combine data from cameras, LiDAR, radar, and other sensors to create a comprehensive understanding of the environment. Furthermore, reinforcement learning and probabilistic roadmap planning algorithms contribute to complex decision-making and route planning in L3-L4 AVs. L5 (full automation) requires robust sensor fusion, 3D mapping capabilities, and deep reinforcement learning approaches for adaptive behavior prediction and high-level route planning.\\

Some industry relevant examples have been illustrated below:\\ 
\textbf{Kodiak}
\begin{itemize}
    \item Status: Kodiak currently operates a fleet of Level 4 autonomous trucks for commercial freight hauling on behalf of shippers.
    \item Recent Developments:
\begin{itemize}
        \item Kodiak is focusing on scaling its autonomous trucking service as a model, providing the driving system to existing carriers.
        \item The company recently secured additional funding to expand its operations and partnerships.
        \item No immediate news about deployment of driverless trucks beyond current operations.
\end{itemize}

\end{itemize}

\textbf{Waymo}
\begin{itemize}
    \item Status: Waymo remains focused on Level 4 autonomous vehicle technology, primarily targeting robotaxi services in specific geographies.
    \item Recent Developments:
\begin{itemize}
        \item Waymo is expanding its robotaxi service in Phoenix, Arizona, with plans to eventually launch fully driverless operations.
        \item The company's Waymo Via trucking division continues testing autonomous trucks in California and Texas.
        \item No publicly announced timeline for nationwide deployment of driverless trucks.
\end{itemize}

\end{itemize}

\textbf{Overall:}
\begin{itemize}
    \item Both Kodiak and Waymo are making progress towards commercializing Level 4 autonomous vehicles, but primarily focused on different segments (trucks vs. passenger cars).
    \item Driverless truck deployment timelines remain flexible and dependent on regulatory approvals and further testing as was discussed previously.
\end{itemize}

\paragraph{\textbf{Key AI/Learning Components across Levels}}
\begin{itemize}
\item Perception:
\begin{itemize}
    \item L0-L2: Basic object detection and lane segmentation using CNNs.
    \item L3-L4: LiDAR-based object detection, advanced sensor fusion algorithms for robust object recognition.
    \item L5: 3D object mapping, robust sensor fusion and interpretation.
\end{itemize}
\item Decision-Making:
\begin{itemize}
    \item L0-L2: Rule-based algorithms for lane change assistance, adaptive cruise control.
    \item L3-L4: Probabilistic roadmap planning (PRM), decision-making models for route selection.
    \item L5: Deep reinforcement learning for adaptive behavior prediction, high-level route planning.
\end{itemize}
\item Control:
    \begin{itemize}
        \item L0-L2: PID controllers [\cite{22}] for basic acceleration and braking adjustments.
        \item L3-L4: Model Predictive Control (MPC) [\cite{35}] for complex maneuvers, trajectory tracking algorithms.
        \item L5: Multi-task DNNs for real-time coordination of all driving actions.
    \end{itemize}
\end{itemize}
The following Table \ref{tab:levels_table} provides examples of AI algorithms used at different autonomy levels, from L0 to L5, highlighting key techniques and applications. We have considered the percentage of systems using AI algorithms, algorithm types, examples of AI algorithms at each level and the key tasks being automated at each level of autonomy. Please note that at L0, the extent to which AI or learning algorithms being used is very minimal and not complete algorithms in themselves, although there could be some partial techniques being used like data processing or detecting an object on road

\begin{table}[htp]
    \centering
     \fontsize{6.5pt}{7.5pt}\selectfont
\caption{Statistics on AI and Learning Algorithms in Autonomous vehicles based on Levels of Automation}
\label{tab:levels_table}
    \begin{tabular}{|b{0.8cm}|b{1.0cm}|b{1.3cm}|b{1.2cm}|b{1.0cm}|b{1.0cm}|} \hline 
         \textbf{Level of Autonomy}&  \textbf{\% of Systems Using AI/Learning Algorithms}&  \textbf{Algorithm Types}& \textbf{ Key AI/Learning Algorithms}&  \textbf{Key Tasks Automated}& \textbf{Number of AI/Learning Algorithms} \\ \hline 
 L0 (No Automation)& 0\%& N/A& N/A& N/A&0 \\\hline
         L1 (Driver Assistance)&  50-70\%&  Rule-based systems, Decision trees, Naive Bayes&  Adaptive Cruise Control, Lane Departure Warning (LDW), Automatic Emergency Braking (AEB)&  Sensing, basic alerts and interventions& 3-5 \\ \hline 
         L2 (Partial Automation)&  80-90\%&  Rule-based systems, Decision trees, Reinforcement learning (RL), Support Vector Machines (SVM)&  Traffic Sign Recognition, Highway Autopilot (ACC + lane centering), Traffic Jam Assist&  Navigation, lane control, stop-and-go, limited environmental adaptation& 5-10 \\ \hline 
         L3 (Conditional Automation)&  90-95\%&  Deep Learning (DL) (e.g., Convolutional Neural Networks, Recurrent Neural Networks), RL, Probabilistic models&  Urban Autopilot, Valet Parking&  Full control under specific conditions, dynamic environment adaptation, complex decision-making& 10-15 \\ \hline 
         L4 (High Automation)&  95-99\%&  Advanced DL (e.g., Generative Adversarial Networks, Transformers), Multi-agent RL, Sensor fusion algorithms&  City Navigation, Highway Chauffeur&  Full control in specific environments, high-level navigation, complex traffic scenarios& 15-20 \\ \hline
 L5 (Full Automation)& 100\%& Advanced DL, Multi-agent RL, Hybrid algorithms (combining various types), Explainable AI (XAI)& Universal Autonomy

& Full control in all environments, self-learning and adaptation, human-like decision-making&20+ \\\hline
    \end{tabular}
\end{table}

The level of autonomy in an AV directly correlates with the size of its software package. Imagine a pyramid, with Level 0 at the base (smallest size) and Level 5 at the peak (largest size). Each level adds functionalities and complexities, reflected in the increasing size of the pyramid.

Challenges and Implications:
 \begin{itemize}
     \item Limited Storage \& Processing Power: Current onboard storage and processing capabilities might not yet be sufficient for larger Level 4 and 5 software packages.
     \item Download and Update Challenges: Updating these larger packages may require longer download times and potentially disrupt vehicle operation.
     \item Security Concerns: The more complex the software, the higher the potential vulnerability to cyberattacks, necessitating robust security measures.
 \end{itemize}

AV software package size is a major challenge for developers like Nvidia and Qualcomm, as larger packages require:

\begin{itemize}
    \item Increased processing power and memory: This translates to higher hardware costs and potentially bulkier systems.
    \item Slower download and installation times: This can be frustrating for users, especially in areas with limited internet connectivity.
    \item Security concerns: Larger packages offer more attack vectors for potential hackers.
\end{itemize}

Here's how Nvidia and Qualcomm are tackling this challenge:

\textbf{Nvidia:}

\begin{itemize}
    \item Drive Orin platform: Designed for high-performance AV applications, Orin features a scalable architecture that can handle large software packages.
    \item Software optimization techniques: Nvidia uses various techniques like code compression and hardware-specific optimizations to reduce software size without sacrificing performance.
    \item Cloud-based solutions: Offloading some processing and data storage to the cloud can reduce the size of the onboard software package.[\cite{15}]
\end{itemize}

\textbf{Qualcomm:}

\begin{itemize}
    \item Snapdragon Ride platform: Similar to Orin, Snapdragon Ride is a scalable platform built for efficient processing of large AV software packages.
    \item Heterogeneous computing: Qualcomm utilizes different processing units like CPUs, GPUs, and NPUs to optimize performance and reduce software size by distributing tasks efficiently.
    \item Modular software architecture: Breaking down the software into smaller, modular components allows for easier updates and reduces the overall package size. [\cite{16}]
\end{itemize}

\textbf{Additional approaches:}

\begin{itemize}
    \item Standardization: Industry-wide standards for AV software can help reduce duplication and fragmentation, leading to smaller package sizes.
    \item Compression algorithms: Advanced compression algorithms can significantly reduce the size of data and code without compromising functionality.
    \item Machine learning: Using machine learning to optimize software performance and resource utilization can help reduce the overall software footprint.
\end{itemize}
The battle against AV software package size is ongoing, and both Nvidia and Qualcomm are at the forefront of developing innovative solutions. As technology advances and these approaches mature, we can expect to see smaller, more efficient AV software packages that pave the way for wider adoption of self-driving vehicles.

Here's a deeper dive as shown in the Table \ref{tab:SW_pack_size} into this relationship[\cite{36}].

\begin{table}[h!]
    \centering
\caption{Software Package Size based on the Levels of Autonomy}
\label{tab:SW_pack_size}
    \begin{tabular}{|c|l|} \hline 
          Level of Autonomy&Package Size\\ \hline 
          0&Few MB\\ \hline 
          1&100s of MB\\ \hline 
          2&100s MB to Few GBs\\ \hline 
          3&Few GB to 10's of GB\\ \hline 
          4&10's of GB to 100's of GB\\ \hline
 5& 100's of GB to TBs\\\hline
    \end{tabular}

\end{table}
The level of autonomy directly influences the size of an AV's software package. While higher levels offer greater convenience and potential safety benefits, they come with the challenge of managing increasingly complex and computationally intensive software packages that would require large storage spaces that the current processors cannot accommodate. Hence, the transformation towards zonal-based architectures is desirable with multiple but small number of processors that are tasked to accomplish a particular function or task providing the ample of storage space needed for moving towards Level 5 along with supporting connected and automated vehicle concepts.

\section{Conclusion} 

This paper presents a comprehensive analysis of the role of AI and learning algorithms in autonomous vehicles, diving into various aspects such as the shifts from rule-based systems to deep neural networks due to improved model capabilities and computing power. The specific needs for trucks vs. cars have been detailed like the trucks prioritizing hub-to-hub transport and efficient long-haul journeys, while passenger cars aiming for source-to-destination autonomy. Increasing complexity from basic object detection (L0-L2) to 3D mapping and adaptive behavior prediction (L3-L5) has been explained. Challenges and implications such as limited storage and processing power, software update concerns, and increased security vulnerabilities at higher autonomy levels have been discussed. Some of the key conclusions include that AI is crucial for achieving different levels of autonomous vehicle functionality. Advanced techniques like deep learning and reinforcement learning are essential for higher levels with complex decision-making and adaptable behavior.
Truck and car AI applications have distinct requirements. Trucks focus on route optimization and fuel efficiency, while passenger cars prioritize passenger comfort and dynamic adaptation to urban environments.
Software package size grows with autonomy level. This poses challenges for storage, processing, and software updates, highlighting the need for efficient architectures and robust security measures.
Further research is needed on self-driving trucks despite their promising business potential. This area lags behind passenger car research, but its development could significantly optimize logistics and address driver shortages.

Certain aspects fall outside the scope of this paper, including emerging technologies and AI algorithms like quantum AI, transfer learning, and Meta-Learning. Currently, these applications in robotics and physical systems, such as vehicles, are limited. However, the paper does not rule out their potential use within vehicles for tasks like edge computing or learning human behaviors through transfer learning. In addition, the study on trends in published research in artificial intelligence (Stat 1) only considers the four most popular platforms within a broader automotive industry. The paper acknowledges the existence of other journals and conferences, such as NeurIPS [\cite{37}], which may have more publications. A quick study on these new platforms revealed similar trends to those presented in the paper. The exclusion of discussions on emerging trends is intentional to narrow the scope and present relevant studies for drawing meaningful conclusions.
The paper concludes by presenting a clear image of the evolving AI landscape in autonomous vehicles, stressing its critical role in efficient and safe transportation solutions. It identifies key challenges and suggests areas for future research, contributing to a road map for researchers, practitioners, and enthusiasts interested in the dynamic relationship between AI, learning algorithms, and the forefront of contemporary transportation.


%


\ifCLASSOPTIONcaptionsoff
  \newpage
\fi



%

%

\begin{IEEEbiography}[{\includegraphics[width=1in,height=1.25in,clip,keepaspectratio]{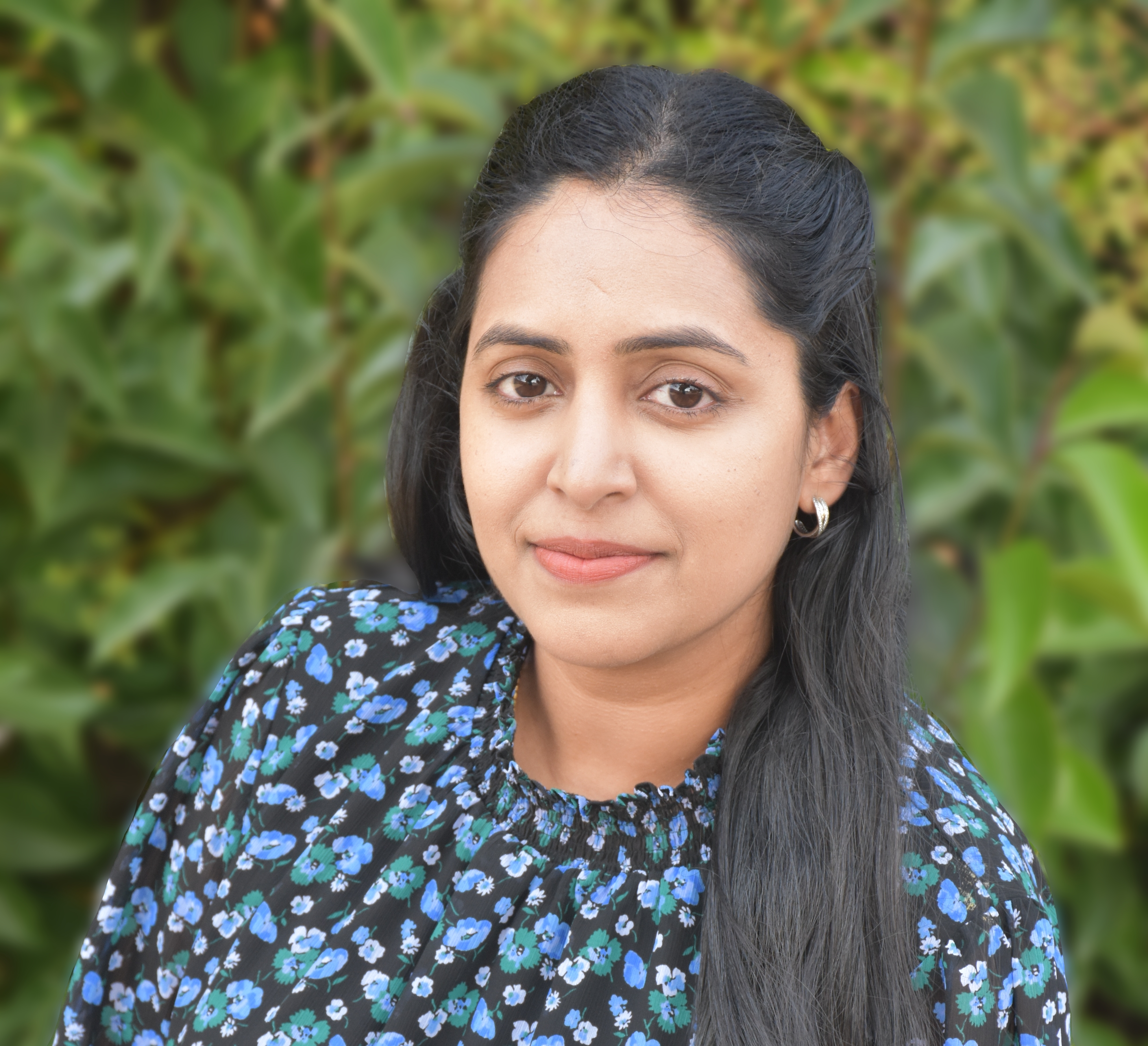}}]{Divya Garikapati}is a Senior IEEE Member and currently serving as the standards committee member within the IEEE Intelligent Transportation Systems Society (ITSS) and a peer reviewer for IEEE Intelligent Transportation Systems Conferences (ITSC). She actively participates in several industry level standards discussions within IEEE and SAE organizations. She is also the working group chair for the IEEE Vehicular Technology Society standards discussions. She received her Masters in Electrical Engineering Systems from the University of Michigan, Ann Arbor in 2014. Prior to that, she received her Bachelors in Electronics and Communications Engineering from Andhra University College of Engineering, Andhra Pradesh, India. Her current work focuses on Systems and Safety research and development for Level 2,3 and 4 Autonomous vehicles. She has over 10 years of experience in the automotive industry.
\end{IEEEbiography}
\begin{IEEEbiography}[{\includegraphics[width=1in,height=1.25in,clip,keepaspectratio]{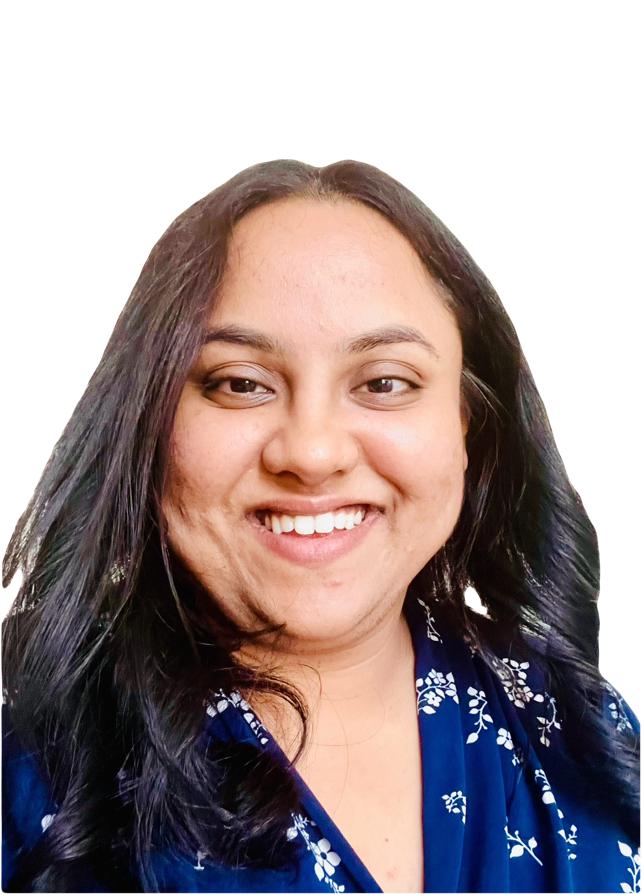}}]{Sneha Sudhir Shetiya}is a Senior IEEE Member and received her Maters degree in Electrical Engineering with a major in computer vision and Signal Processing from North Carolina State University, in 2021. She received her Bachelor's degree in Electronics and Communication Engineering from the Visvesvaraya Technological University, Karnataka, India, in 2014. Her work involves middleware topics for embedded development of autonomous driving stack, automotive diagnostics, systems engineering and functional safety. She is an active volunteer with IEEE region 4 and takes part in activities of Women In Engineering (WIE) groups in the region. She is part of the committee for senior member evaluation at IEEE for 2024 and has been a proctor for IEEExtreme 24 hour coding competition.
\end{IEEEbiography}
 \vfill{}
 \vfill{}

\end{document}